\documentclass[runningheads]{llncs}

\usepackage{eccv}

\usepackage{eccvabbrv}

\usepackage{graphicx}
\usepackage{booktabs}
\usepackage{multirow}
\usepackage[accsupp]{axessibility}  

\usepackage{soul}

\usepackage{hyperref}

\usepackage{orcidlink}

\begin{document}
\renewcommand{\arraystretch}{0.85}
\title{CycleBNN: Cyclic Precision Training in Binary Neural Networks} 

\titlerunning{CycleBNN: Cyclic Precision Training in Binary Neural Networks}

\author{Federico Fontana\inst{1}\orcidlink{0009-0007-0437-7832} \and
Romeo Lanzino\inst{1}\orcidlink{0000-0003-2939-3007} \and
Anxhelo Diko\inst{1}\orcidlink{0000-0002-3084-9377}
\and
Gian Luca Foresti\inst{2}\orcidlink{0000-0002-8425-6892}
\and
Luigi Cinque\inst{1}\orcidlink{0000-0001-9149-2175}
}

\authorrunning{F. Fontana et al.}

\institute{Sapienza University of Rome, Rome RM 00198, Italy \\ \email{ \{fontana.f, lanzino, diko, cinque\}@di.uniroma1.it} \and
UNIUD University of Udine,  Udine UD 33100, Italy \\
\email{ gianluca.foresti@uniud.it}}

\maketitle

\begin{abstract} This paper works on Binary Neural Networks (BNNs), a promising avenue for efficient deep learning, offering significant reductions in computational overhead and memory footprint to full precision networks. However, the challenge of energy-intensive training and the drop in performance have been persistent issues. Tackling the challenge, prior works focus primarily on task-related inference optimization. Unlike prior works, this study offers an innovative methodology integrating BNNs with cyclic precision training, introducing the CycleBNN. This approach is designed to enhance training efficiency while minimizing the loss in performance. By dynamically adjusting precision in cycles, we achieve a convenient trade-off between training efficiency and model performance. This emphasizes the potential of our method in energy-constrained training scenarios, where data is collected onboard and paves the way for sustainable and efficient deep learning architectures.
To gather insights on CycleBNN's efficiency, we conduct experiments on ImageNet, CIFAR-10, and PASCAL-VOC, obtaining competitive performances while using 96.09\% less operations during training on ImageNet, 88.88\% on CIFAR-10 and 96.09\% on PASCAL-VOC. Finally, CycleBNN offers a path towards faster, more accessible training of efficient networks, accelerating the development of practical applications.
The PyTorch code is available at  \url{https://github.com/fedeloper/CycleBNN/}.
\end{abstract}

\section{Introduction}
\label{sec:intro}
\begin{figure}[ht]
    \centering
    \begin{minipage}{0.35\textwidth}
        \centering
        \includegraphics[width=\linewidth]{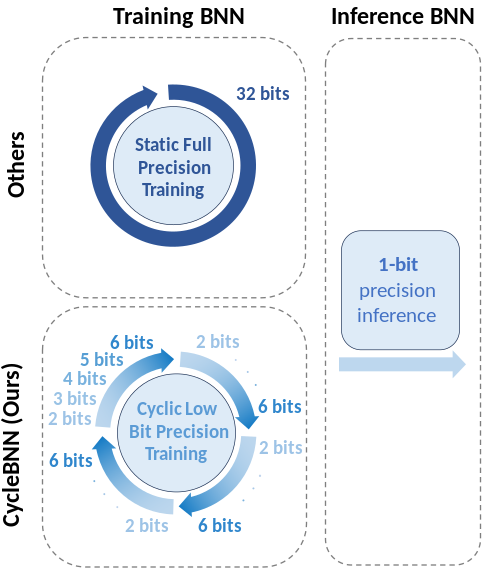} 
        \caption*{(a)}

    \end{minipage}\hfill
    \begin{minipage}{0.65\textwidth}
        \centering
         \includegraphics[width=\linewidth]{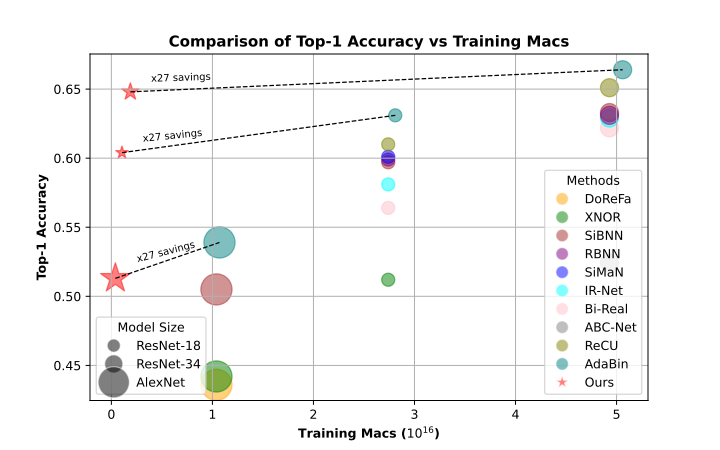}
        
        \caption*{(b)}
        \label{comparison}
    
    \end{minipage}
     \caption{(a) Illustrates the standard pipeline for training and inference in BNN alongside the proposed pipeline incorporating CycleBNN.
     (b) Compares the accuracy/training MACs across various models and architectures on ImageNet.}\label{system_overview}
\end{figure}

In the current era of deep learning, Deep Neural Networks (DNNs) have made revolutionary strides in a variety of domains, from computer vision~\cite{krizhevsky2012imagenet} to natural language processing~\cite{bahdanau2014neural}, and beyond. This has given rise to more complex network architectures for superior model performance. However, the complexity of these networks also introduces challenges in computational overhead, energy consumption, and storage requirements, particularly when deployment is targeted at resource-constrained environments.
Different approaches for enhancing network efficiency can be categorized into quantization and pruning \cite{choquette2020nvidia,he2017channel,li2016pruning,10438214}.
One of the significant innovations in this space is the introduction of Binary Neural Networks(BNNs). BNNs are DNNs with both weights and activations using only 1-bit values, which offer a paradigm shift in memory usage and computational requirements as compared to their full-precision counterparts~\cite{han2020training,rastegari2016xnor,xu2021learning,yang2020searching}. This binarization replaces multiplication and accumulation operations with more efficient bit-wise operations like XNOR and BitCount. The resulting networks significantly increased computational speed and reduced memory requirements during inference ~\cite{rastegari2016xnor}.
Nevertheless, the exploration of BNNs has encountered its challenges. Although highly efficient, BNN models experience significant degradation in accuracy compared to full-precision models. Furthermore, although inference in BNN is more efficient than a full-precision network, the training process does not achieve similar efficiency gains. This is because the training uses full precision (32-bit representation). This drop in accuracy stems from the inherent limitations of 1-bit representations~\cite{30711}.

The existing body of literature on BNNs primarily aims to enhance network performance \cite{rastegari2016xnor,lin2017towards,qin2020forward,bethge2019binarydensenet,tu2022adabin} or extend their applicability to additional tasks. Conversely, this study emphasizes improving training efficiency, delving into a novel research domain for BNNs \cite{wang2020bidet,10.1007/978-3-031-27077-2_8,wei2023ebsr,10049753,qin2021bipointnet,lanzino2024faster}. Few attempts at efficient training have been made for BNNs, but these were either applied to toy datasets or yielded limited efficiency gain \cite{laydevant2021training,wang2021enabling}.

In response, this research presents CycleBNN, a methodology that combines cyclic precision training with BNNs, fundamentally based on dynamically adjusting precision throughout the training process. This means that the precision with which weights and activations are represented changes periodically throughout training to optimize performance and efficiency. This approach aims to achieve an optimal equilibrium between training efficiency and model efficacy.
Figure \ref{system_overview} illustrates the pipeline of a standard BNN and contrasts it with CycleBNN, graphically highlighting the concept of cyclical quantization.

We conducted several experiments to evaluate CycleBNN on ImageNet, CIFAR-10, and PASCAL-VOC to evaluate the method. Notably, while achieving competitive performance with state-of-the-art techniques, CycleBNN stands out with 96.09\% less operations during training on ImageNet, 88.88\%  on CIFAR-10, and 96.09\%  on PASCAL-VOC. The implications of this work are profound, especially in scenarios with energy constraints or where data collection and processing occur onboard. Methods like CycleBNN offer a promising glimpse into sustainable and scalable solutions that can cater to various applications, from edge devices to large-scale data centers.

This work's contributions can be summarized as follows:
\begin{itemize}
    \item We rethink the training of BNNs, proposing an integration of cyclic precision scheduling called CycleBNN, offering a dynamic adaptation of model precision during training. Furthermore, we formalize the math behind the training of this network and its cyclic precision.
    \item Extensive evaluation of CycleBNN's performance on two distinct tasks, image classification and object detection, showcasing a negligible trade-off between training efficiency and performance.
    \item An extensive ablation study on hyper-parameters and a commitment to open science by releasing the code and all necessary components to ensure reproducibility of the reported results, encouraging further research and development in efficient deep learning architectures.
\end{itemize}

\section{Related Work}
The concept of a BNN was pioneered by~\cite{courbariaux2016binarized}. Their innovative approach entailed using the sign function to binarize weights and activations. This method effectively replaced the majority of arithmetic computations in deep neural networks with bit-wise operations.

BNNs are impacted by quantization error due to the reduction of weight precision to binary values.
To address this issue, XOR-Net~\cite{rastegari2016xnor} introduced a channel-wise scaling factor to reconstruct binarized weights. This technique became a crucial component in many later BNN models. ABC-Net, proposed by~\cite{lin2017towards}, aimed to approximate the full-precision weights by utilizing a linear combination of several binary weight bases. They also incorporated multiple binary activations to reduce information loss.

Drawing inspiration from the architecture of ResNet~\cite{he2016deep} and DenseNet~\cite{huang2017densely}, Bi-Real Net~\cite{liu2018bi} introduced shortcuts to narrow the performance discrepancy between 1-bit and real-valued CNN models. Meanwhile, BinaryDenseNet~\cite{bethge2019back} augmented the accuracy of BNNs by expanding the number of shortcuts.

Additionally, IR-Net~\cite{qin2020forward} introduced the Libra-PB method, which aims to minimize information loss during forward propagation. This is achieved by maximizing the information entropy of quantized parameters and curbing the quantization error within the constraint of $\left\{-1, +1\right\}$. Furthermore, ReActNet~\cite{liu2020reactnet} presented a generalized version of the traditional sign and PReLU functions, termed RSign and RPReLU, respectively. These generalized functions enable explicit learning of distribution reshaping and shifting with negligible additional computational expense.
RBNN \cite{lin2020rotated} investigates and mitigates the impact of angular bias on quantization error. SiMaN \cite{lin2021siman} demonstrates that eliminating $L_2$ Regularization throughout the training process maximizes entropy. ReCU \cite{xu2021recu} proposes a rectified clamp unit designed to rejuvenate "dead weights," thereby reducing the quantization error.
AdaBin~\cite{tu2022adabin} incorporates equalization techniques for weights and introduces learnable parameters for activations. It also employs the MaxOut nonlinear activation function, adding a minimal number of floating-point operations. BiDet \cite{wang2020bidet} introduces an innovative framework for efficient object detection using binarized neural networks. Following this, AdaBin \cite{tu2022adabin}, advances the field further by enhancing this technique for object detection.

Several attempts to train DNNs with low precision have gained significant interest due to their efficacy in enhancing the time and energy efficiency of the training process \cite{fu2021cpt,kim2022power,yu2022ldp}. Experimental results and visual analyses corroborate that this cyclic precision training approach not only facilitates convergence to broader minima, leading to reduced generalization errors, but also diminishes training variance \cite{fu2021cpt}. This finding suggests the potential for a novel avenue in enhancing the optimization and efficiency of DNN training simultaneously.

\section{Cyclic Precision in Binary Neural Networks}

This section delves into the process of binarizing both weights and activations in BNNs, alongside an exploration of cyclic precision's role during training. Initially, it introduces BNNs, detailing the binarization process and convolution adaptation. Following this, the method of weight equalization is outlined. Subsequently, the concept of cyclic precision is introduced, integrating seamlessly with BNNs training pipeline for enhanced training efficacy.
\subsection{Binary Convolution}
In this subsection, we describe how the convolution operation changes in the BNNs context and how the gradient is approximated.

For a given input tensor $a \in \mathbb{R}^{c_{in}\times h\times w}$ and a weight tensor of a 2D convolution $w \in \mathbb{R}^{c_{in}\times c_{out}\times k_h\times k_w}$, the output after the $\mathit{Conv}$ operation is given by:

\begin{equation}
\textit{out} = \mathit{Conv}(a, w)
\end{equation}

The output, denoted as $\textit{out} \in \mathbb{R}^{c_{out}\times h'\times w'}$, is shaped by the stride, padding, filter dimensions, and the number of channels. Here, $h'$ and $w'$ signify the output height and width, respectively. The parameters \(h\), \(w\), \(c_{in}\), \(c_{out}\), \(k_h\), and \(k_w\) specify the input image's height, width, input channels, output channels, kernel height, and kernel width, respectively. Moreover, all computational operations are executed with full precision, utilizing a 32-bit representation.

In the context of BNNs, the binarization of the activation and the weight tensors is a crucial step. This is achieved by representing every value within these tensors as either $-1$ or $+1$. The decision criterion is decided by the $\mathit{Sign}$ function:

\begin{equation}
\mathit{Sign}(x)=\left\{
\begin{aligned}
-1, & \quad x<0 \\
+1, & \quad x\ge0 \\
\end{aligned}
\right. 
\label{binarization}
\end{equation}
In this context, the floating-point multiplications are replaced by bit-wise operations to accelerate the computations. Specifically, the \textit{XOR} operation is utilized, followed by a \textit{BitCount} operation.

For two binary tensors $a_b \in \{-1,1\}^{c_{in}\times h\times w}$ and $w_b \in \{-1,1\}^{c_{in}\times c_{out}\times k_h\times k_w}$, the output of the operation is:

\begin{equation}
\textit{out}_b = \mathit{BitCount}(a_b \oplus \textrm{w}_b)
\end{equation}

where $\textit{out}_b \in \mathbb{N}^{c_{out}\times h'\times w'}$ and $\oplus$ denotes the \textit{XOR} operation. The 
\textit{BitCount} function counts the number of bits with a value of 1 for each element in the resultant tensor from the \textit{XOR} operation. The resultant computational paradigm reduces the overhead and the latency and energy used, making the entire process considerably more efficient\cite{rastegari2016xnor}.

To train a BNN, the binarization of weights \(w_b\) and activations \(a_b\) during the forward pass is performed according to Eq.~(\ref{binarization}). In contrast, the real-valued weights \(w_r\) and activations \(a_r\) are updated during the backpropagation phase. However, the derivative of \( \textit{Sign}(x) \) is 0 because the function is constant (either -1 or 1), which poses challenges for the optimization process. Therefore, we choose to implement the Straight-Through Estimator (STE)~\cite{Bengio2013EstimatingOP} to approximate the gradient.

\subsection{Equalization of Weights}
The process of low-bit quantization significantly diminishes the ability of filter weights to extract features, and this effect is even more pronounced for the 1-bit scenario. A range of methods have been employed in prior BNNs to optimize these binarized weights. 
XNOR-Net~\cite{rastegari2016xnor} addresses this by reducing the mean squared error (MSE) through the introduction of a scaling factor. Similarly, IR-Net~\cite{qin2020forward} maximizes the information entropy by reshaping the weights, applying a method analogous to that of XNOR-Net. AdaBin \cite{tu2022adabin} introduce Kullback-Leibler divergence~\cite{kullback1951information} to evaluate the information loss.
ReCU \cite{xu2021recu} proposes to use the standardization, but not the centralization noting that most of the time the mean is equal to 0.
In this work, as~\cite{Qin2020ForwardAB}, we maximize the information entropy by standardizing the weights in each forward propagation as follows
\begin{equation}
    \mathcal{W}^\prime = \dfrac{\mathcal{W}}{\sigma (\mathcal{W})}
    \label{eq:w_hat}
\end{equation}
where $\sigma(\cdot)$   denotes the standard deviation. 

\subsection{Cyclic Precision during Training}
In this subsection, we first show the formula for static quantization and then define the concept of quantization error and the rationale behind training BNNs with very low bits precision. Finally, we introduce the cyclic precision scheduling equation. We conclude by introducing the cyclic precision scheduling equation and presenting the equations for gradient approximation concerning weights and activations via the STE.

Precision in DNNs, denoted as \( p \), refers to the number of bits used to represent a number. The quantization process at \( p \) bits reduces the range of possible values from a broad spectrum to a more confined set. This reduction is crucial for optimizing the use of computational resources and memory, as it allows for more efficient management by limiting the precision of the data. For an \( p \)-bit quantization, the number of possible quantized values amounts to \( 2^p -1 \). To simulate a precision of \( p \) bits, the static quantization function \( Q(x,p,m,M) \) is defined as:
\begin{equation}
Q(x,p,m,M) = m + \frac{\left\lfloor \frac{(x-m)(2^p-1)}{M-m}+\frac{1}{2} \right\rfloor (M-m)}{2^p -1}
\end{equation}
where \( \left\lfloor \cdot \right\rfloor \) is the \textit{Floor} function, $M$ and $m$ are the maximum and minimum values, respectively.

The quantization error is defined as:
\begin{equation}
    \operatorname{QE}(w_{32}) = \int_{-\infty}^{+\infty}f(w_{32})\big(w_{32} - \alpha \textit{Sign}(w_{32})\big)^2\mathrm{d}w_{32}
\end{equation}
where \( f(w_{32}) \) is the probability density function of \( w_{32} \), and \( \alpha \) is a scaling factor. 
Throughout the paper, we utilize subscripts to denote the application of \(Q\) with \(p\)  set as the subscript value, applying this notation consistently in the presented equation and subsequent discussions. 
The term \( \alpha \mathit{Sign}(w_{32}) \) in the equation represents binarization, where \( \textit{Sign}(w_{32}) \) maps \( w_{32} \) to \(\{-1, 1\}\) based on its sign, and \( \alpha \) scales it. When \( w_{32} \) is quantized to a lower dimension, the range of possible values for \( w_{32} \) decreases, leading to a reduction in the variability of \( w_{32} \). This reduction generally results in a decrease in the term \( (w_{32} - \alpha \textit{Sign}(w_{32}))^2 \) for each \( w_{32} \), as the quantized value \( \alpha \textit{Sign}(w_{32}) \) becomes closer to the original \( w_{32} \) value. Therefore, the integral, which sums up these squared differences weighted by the probability density \( f(w_{32}) \), is likely to be smaller when the dimensionality of quantization is lower. The overall distance between the original and quantized weights is reduced across the distribution of \( w_{32} \).\footnote{Evidence supporting these claims is provided in the supplementary material.}

The cyclic precision technique adjusts the bit precision based on the current epoch. We define the Precision Scheduling function (\( \textit{PS} \)) as follows:
\begin{equation}
\textit{PS}(e,N,c,v,V) = \left\lfloor \left( v + \frac{(V + 1 - v) e c}{N} \right) \bmod (V + 1 - v) \right\rfloor\label{scheduling}
\end{equation}
where \( e \) is the current training epoch, \( N \) is  the total number of epochs, \( c \) denotes the number of precision cycles, and \( V  \) and \( v \) are the maximum and minimum bit precisions, respectively.

The process outlined in Equations \ref{g1} and \ref{g2} involves the binarization of weights \(w_b\) and activations \(a_b\) according to Equation~(\ref{binarization}) during the forward pass. Gradient approximation is conducted using precision \textit{PS} for all terms except for \(\mathcal{L}\), which is computed with 8-bit precision. This methodology enables the updating of real-valued weights \(w_{32}\) and activations \(a_{32}\) with enhanced accuracy.

To approximate the gradient for weights using STE and precision scheduling, the formula is given by:
\begin{equation}
\frac{\partial \mathcal{L}_{32}}{\partial w_{32}} = \frac{\partial \mathcal{L}_{32}}{\partial w_{1}} \cdot \frac{\partial w_{1}}{\partial w_{32}} \approx \frac{\partial \mathcal{L}_{8}}{\partial w_{1}}
\label{g1}
\end{equation}

where the gradient approximation for the activations is described through:
\begin{equation}
\frac{\partial \mathcal{L}_{32}}{\partial a_{32}} = \frac{\partial \mathcal{L}_{32}}{\partial a_{1}} \cdot \frac{\partial a_{1}}{\partial a_{32}} \approx \frac{\partial \mathcal{L}_{8}}{\partial a_{1}} \cdot \frac{\partial F(a_{PS(e)})}{\partial a_{\textit{PS}(e)}}
\label{g2}
\end{equation}

The piece-wise polynomial function for approximating the gradient with respect to activations is given as:
\begin{equation}
\frac{\partial F(a_{\textit{PS}(e)})}{\partial a_{\textit{PS}(e)}} = \begin{cases}
2 + 2a_{\textit{PS}(e)}, & \text{if } -1 \le a_{\textit{PS}(e)} < 0, \\
2 - 2a_{\textit{PS}(e)}, & \text{if } 0 \le a_{\textit{PS}(e)} < 1, \\
0, & \text{otherwise.}
\end{cases}
\end{equation}
In these equations, \(a\) represents the activations, \(w\) denotes the weights, and \(\mathcal{L}\) is the loss function.

\begin{figure}[t]
	\begin{center}
		\includegraphics[width=0.60\linewidth]{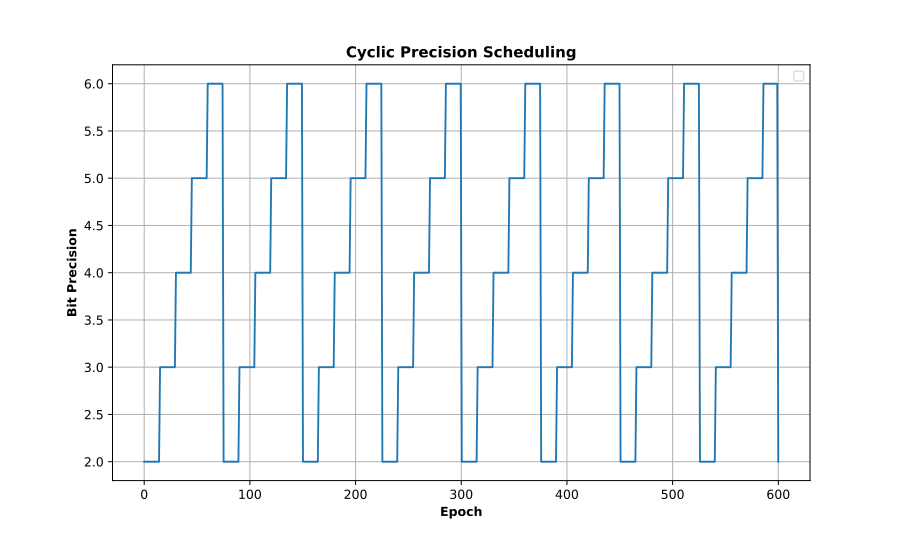}
	\end{center}
	\caption{An example of cyclic precision scheduling, following Eq. \ref{scheduling}, with adjustments between 2-6 bits, across a total of 600 epochs, distributed over 8 cycles}
	\label{prec_sched}
\end{figure}

\begin{figure}[t]
	\begin{center}
		\includegraphics[height=0.5\linewidth]{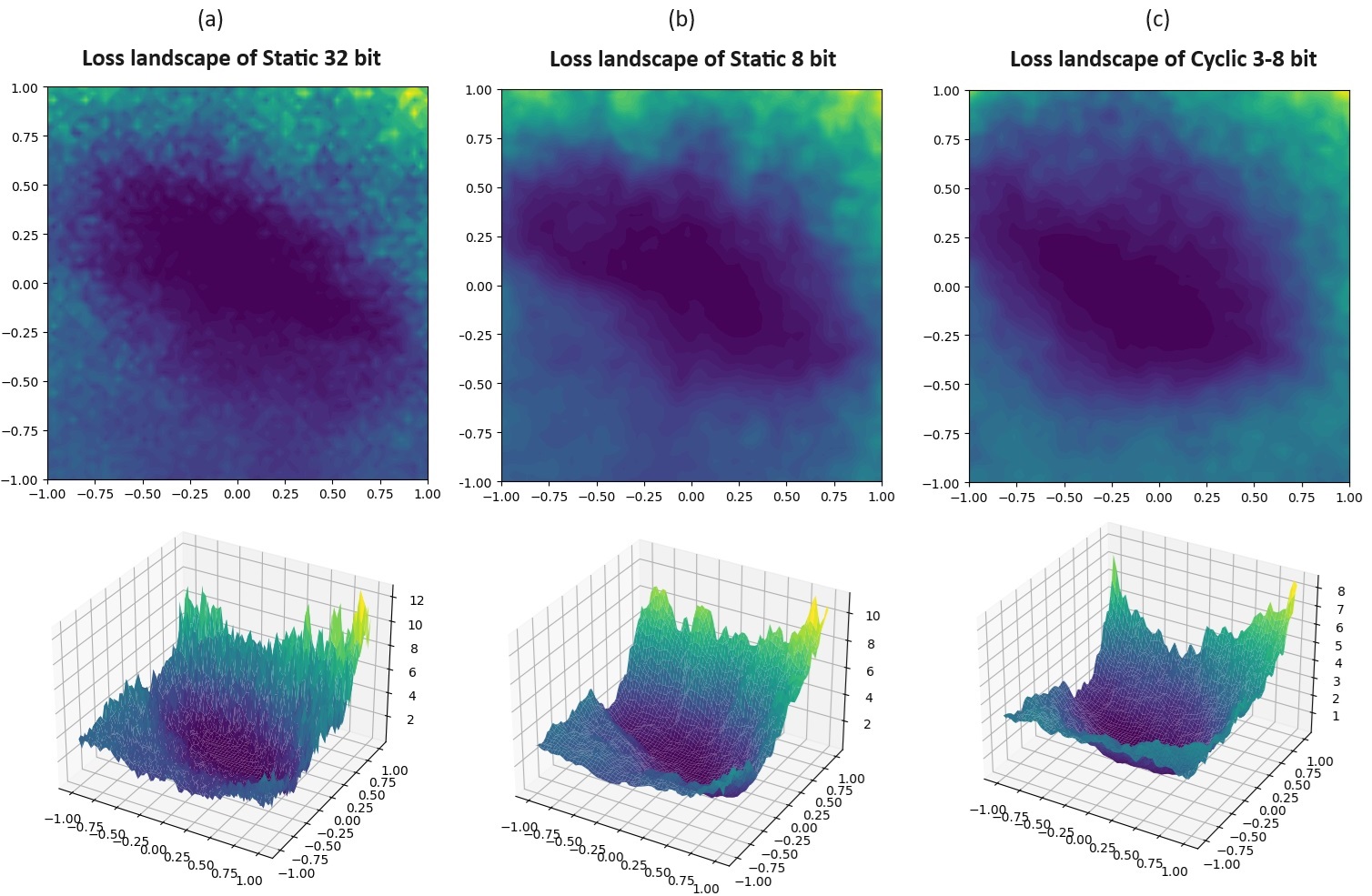}
	\end{center}
	\caption{Loss landscape visualization after convergence of ResNet-18 (with W/A 1 bit) on CIFAR-10 trained with different precision schedules, where wider contours with larger intervals indicate a better local minima and a lower generalization error as analyzed in \cite{li2018visualizing}.}
	\label{loss_landscape}
\end{figure}

Figures \ref{prec_sched} and \ref{loss_landscape} illustrate the concept of precision scheduling and its impact on the training process.

\section{Experiments}
In this section, we detail the evaluation metric, datasets, experimental setup, and the results of CycleBNN. We demonstrate the effectiveness of the proposed CycleBNN technique by comparing it with current state-of-the-art methods.

\subsection{Evaluation Metrics} 
We tested the proposed model's performance not only by its accuracy (\textit{Top-1}), but also by cost-sensitive metrics relevant to training expenditure, following \cite{fu2020fractrain,zhou2016dorefa,fu2021cpt,kim2022power,yu2022ldp}. 

To determine the computational expense to train DNNs we use the adjusted count of Multiply-Accumulate operations (MACs)\cite{shen2020fractional}. This adjustment is expressed as:
\begin{equation}
   \textit{Training MACs} = (\text{\# of MACs}) \times (\text{Bit}_{a}/32) \times (\text{Bit}_{b}/32) ,
\end{equation}
 indicating the number of MACs adjusted for the bit precision of operands $a$ and $b$. This metric reflects the total bit operation count, providing a quantifiable measure of computational expense.
 
To express the memory savings from adopting a precision lower than 32, aligning with advancements in \cite{fu2021cpt,kim2022power,yu2022ldp}, we use the \textit{Memory Usage} metric. 
It measures the relative reduction in memory usage by employing lower precision to calculate and store in memory weights, activations, or gradients, compared to the standard 32-bit floating-point precision. It is mathematically represented as:
    \begin{equation}
        \textit{Memory Usage} = \frac{\text{Memory}_{\text{used}}}{\text{Memory}_{\text{32}}} ,
    \end{equation}
    where $\text{Memory}_{\text{used}}$ is the peak memory usage under the precision regime, and $\text{Memory}_{\text{32}}$ is the memory consumption using 32-bit floating-point precision. This metric underscores the efficiency and resource reduction potential of reduced precision formats.

\subsection{Datasets}
In this section, we describe the datasets utilized and their corresponding splits for training and testing.

The CIFAR-10\cite{krizhevsky2009learning} dataset is designed for image classification. It comprises 60000 color images of 32x32 pixels divided into 10 classes. It includes 50,000 training images and 10,000 test images. 

ImageNet \cite{imagenet_cvpr09} is a comprehensive database for image classification. The dataset used in the ILSVRC 2012 challenge, and in this work, includes around 1,200,000 training images, each of 224x224 pixels, 50,000 validation images, and 150,000 testing images across 1,000 classes.

The PASCAL-VOC dataset \cite{Everingham15} comprises images spanning 20 distinct classes. For training purposes, we employed the train-val sets from VOC 2007 and VOC 2012, which encompass approximately 16,000 images together. For evaluation, we utilized the test set from VOC 2007, containing around 5,000 images, consistent with the approaches taken by \cite{tu2022adabin,wang2020bidet}.

\subsection{Experimental Setup}
 
We train the models on CIFAR-10 over 600 epochs utilizing AdamW optimizer \cite{loshchilov2017decoupled} with a momentum setting of 0.9 and a batch size of 16. An initial learning rate of 0.001 was established, later adjusted according to a Cosine Annealing schedule as described in IR-Net \cite{qin2020forward}. The approach to precision scheduling during the training phase involved cyclic adjustments between 2-6 bits with 8 precision cycles. The model's data augmentation and preprocessing were guided by the methodologies specified in \cite{devries2017improved}.

The models were trained using the ImageNet dataset over 120 epochs, employing the AdamW optimizer \cite{loshchilov2017decoupled} with a momentum parameter set to 0.9 and a batch size 16. The training process began with an initial learning rate of 0.001, which was subsequently modified following the Cosine Annealing schedule as outlined in IR-Net \cite{qin2020forward}. Precision scheduling throughout the training was managed through cyclic precision between 2-6 bits across 8 precision cycles. Data augmentation and preprocessing strategies for the model were aligned with the protocols detailed in \cite{he2016deep,tu2022adabin}.

Starting from a pre-trained network on ImageNet, we finetune the model on PASCAL-VOC for 50 epochs, following the indications in~\cite{wang2020bidet}, utilizing Adam\cite{kingma2014adam} with a momentum setting of 0.9 and a batch size of 16. An initial learning rate of 0.01 each 6 epoch it is multiplied by 0.1 until reaching 1e-6. The approach to precision scheduling during the training phase involved cyclic adjustments between 2-6 bits with 8 precision cycles.

 We implemented the proposed method by utilizing the PyTorch\cite{pytorch} framework, and all experiments were run on NVIDIA RTX-4090.
 \subsection{Results}
\noindent\textbf{Results on  CIFAR-10}
The performance of CycleBNN on CIFAR-10 are evaluated against a suite of binary neural network models, including BNN \cite{courbariaux2016binarized}, LAB \cite{hou2016loss}, XNOR-Net \cite{rastegari2016xnor}, DoReFa \cite{zhou2016dorefa}, DSQ \cite{gong2019differentiable}, RAD \cite{ding2019regularizing}, IR-Net \cite{qin2020forward}, RBNN \cite{lin2020rotated}, ReCU \cite{xu2021recu}, AdaBin \cite{tu2022adabin}, and SLB \cite{yang2020searching}. As presented in Table \ref{cifar10_results}, CycleBNN achieved an accuracy of 93.0\% using the ResNet-18 architecture and recorded a 92.1\% accuracy upon binarizing both the weights and activations of the VGG-small architecture into 1-bit representations.

\noindent\textbf{Results on ImageNet}
Subsequently, we confront the performance of CycleBNN making a comparison with XNOR-Net \cite{rastegari2016xnor}, RAD \cite{ding2019regularizing}, IR-Net \cite{qin2020forward}, RBNN \cite{lin2020rotated}, ReCU \cite{xu2021recu}, AdaBin \cite{tu2022adabin}, SLB \cite{yang2020searching} and RBNN \cite{lin2020rotated}. As presented in Table \ref{imagenet_results},
CycleBNN attained accuracies of 51.3\% on AlexNet, 60.4\% on ResNet-18, and 64.8\% on ResNet-50. 
\begin{table}

\caption{Comparisons with state-of-the-art methods on CIFAR-10. W/A denotes the bit width of weights and activations. Training MACs indicates the amount of MACs used during the training on CIFAR-10. In this table, the best values in each column are highlighted in bold, and the second-best values are italicized.}
\label{cifar10_results}

\begin{center}
\begin{tabular}{cccccc} 

\toprule
Networks & Methods & W/A & Top-1(\%) & Training MACs & Memory Usage \\ 
\hline
\multirow{8}{*}{ResNet-18} & Full-precision & 32/32 & 94.8 & 1.05E+15 & 1x \\ 
\cline{2-6}
 & RAD~ & \multirow{6}{*}{1/1} & 90.5 & \multirow{4}{*}{1.05E+15} & \multirow{5}{*}{1x} \\
 & IR-Net~ &  & 91.5 &  &  \\
 & RBNN~ &  & 92.2 &  &  \\
 & ReCU~ &  & 92.8 &  &  \\
 & AdaBin~ &  & \textbf{93.1} & 1.08E+15 &  \\
 & \textbf{\textbf{CycleBNN(Ours)}} &  & \textit{93.0} & \textbf{1.20E+14} & \textbf{0.25x} \\ 
\cline{2-6}
 & Efficiency improv. &  &  & -88.88\% & -75\%~ \\ 
\hline
\multirow{11}{*}{VGG-Small} & Full-precision & 32/32 & 94.1 & 8.75E+15 & 1x \\ 
\cline{2-6}
 & LAB~ & \multirow{9}{*}{1/1} & 87.7 & \multirow{7}{*}{8.75E+15} & \multirow{8}{*}{1x} \\
 & XNOR-Net~ &  & 89.8 &  &  \\
 & BNN~ &  & 89.9 &  &  \\
 & RAD~ &  & 90.0 &  &  \\
 & IR-Net &  & 90.4 &  &  \\
 & RBNN~ &  & 91.3 &  &  \\
 & SLB &  & 92.0 &  &  \\
 & AdaBin &  & \textbf{92.3} & 8.98E+15 &  \\
 & \textbf{\textbf{CycleBNN(Ours)}} &  & \textit{92.1} & \textbf{9.99E+14} & \textbf{\textbf{0.25x}} \\ 
\cline{2-6}
 & Efficiency improv. &  &  & -88.88\% & -75\% \\
\bottomrule
\end{tabular}
\end{center}

\end{table}


\begin{table} [!t]
	\setlength{\tabcolsep}{1mm}
	\renewcommand\arraystretch{0.91}
 \caption{Comparison with state-of-the-art methods on ImageNet for AlexNet and ResNets. W/A denotes the bit width of weights and activations. Training MACs indicates the MACs used during the training on ImageNet. In this table, the best values in each column are highlighted in bold, and the second-best values are italicized.}
	\label{imagenet_results}
	\begin{center}
\begin{tabular}{clcccc} 
\toprule
Networks & Methods & W/A & Top-1(\%) & Training MACs & Memory Usage \\ 
\midrule
\multirow{8}{*}{AlexNet} & Full-precision & 32/32 & 56.6 & 1.04E+16 & 1x \\ 
\cmidrule(lr){2-6}
 & BNN~ & \multirow{6}{*}{1/1} & 27.9 & \multirow{4}{*}{1.04E+16} & \multirow{5}{*}{1x} \\
 & DoReFa~ &  & 43.6 &  &  \\
 & XNOR~ &  & 44.2 &  &  \\
 & SiBNN~ &  & 50.5 &  &  \\
 & AdaBin~ &  & \textbf{53.9} & 1.07E+16 &  \\
 & \textbf{CycleBNN(Ours)} &  & \textit{51.3} & \textbf{3.96E+14} & \textbf{0.25x} \\ 
\cmidrule(lr){2-6}
 & Efficiency improv. & \multicolumn{1}{l}{} &  & -96.09\% & -75\% \\ 
\midrule
\multirow{12}{*}{ResNet-18} & Full-precision & 32/32 & 69.6 & 2.74E+16 & 1x \\ 
\cmidrule(lr){2-6}
 & BNN~ & \multirow{10}{*}{1/1} & 42.2 & \multirow{8}{*}{2.74E+16} & \multirow{9}{*}{1x} \\
 & XNOR-Net~ &  & 51.2 &  &  \\
 & Bi-Real~ &  & 56.4 &  &  \\
 & IR-Net~ &  & 58.1 &  &  \\
 & SiBNN~ &  & 59.7 &  &  \\
 & RBNN~ &  & 59.9 &  &  \\
 & SiMaN~ &  & 60.1 &  &  \\
 & ReCU &  & \textit{61.0} &  &  \\
 & AdaBin~ &  & \textbf{63.1} & 2.81E+16 &  \\
 & \textbf{CycleBNN(Ours)} &  & 60.4 & \textbf{1.05E+15} & \textbf{0.25x} \\ 
\cmidrule(lr){2-6}
 & Efficiency improv. & \multicolumn{1}{l}{} &  & -96.09\% & -75\% \\ 
\midrule
\multirow{10}{*}{ResNet-34} & Full-precision & 32/32 & 73.3 & 4.93E+16 & 1x \\ 
\cmidrule(lr){2-6}
 & ABC-Net~ & \multirow{8}{*}{1/1} & 52.4 & \multirow{6}{*}{4.93E+16} & \multirow{7}{*}{1x} \\
 & Bi-Real~ &  & 62.2 &  &  \\
 & IR-Net~ &  & 62.9 &  &  \\
 & SiBNN~ &  & 63.3 &  &  \\
 & RBNN~ &  & 63.1 &  &  \\
 & ReCU &  & \textit{65.1} &  &  \\
 & AdaBin~ &  & \textbf{66.4} & 5.06E+16 &  \\
 & \textbf{CycleBNN(Ours)} &  & 64.8 & \textbf{1.88E+15} & \textbf{0.25x} \\ 
\cmidrule(lr){2-6}
 & Efficiency improv. & \multicolumn{1}{l}{} &  & -96.09\% & -75\% \\
\cmidrule(lr){2-6}
\end{tabular}
	\end{center}
\end{table}
\noindent \textbf{Results on PASCAL VOC}
 The evaluation of the proposed CycleBNN includes comparisons with the standard structure configuration, as defined in BiDet~\cite{wang2020bidet}. This evaluation extends to a comparison with other binary methods such as BNN~\cite{courbariaux2016binarized} and  XNOR-Net~\cite{rastegari2016xnor},  alongside the task-specific binary detectors BiDet~\cite{wang2020bidet}, AutoBiDet~\cite{9319565} and AdaBin~\cite{tu2022adabin}. Additionally, we include results from multi-bit quantization methods like TWN~\cite{li2016ternary} and DoReFa~\cite{zhou2016dorefa}, which utilize 4-bit weights and activations. Unlike recent literature focusing on 1-bit object detection through Knowledge Distillation~\cite{xu2022ida,ZHAO2022239,9577535}, this study emphasizes training efficiency, comparing models excluding the student-teacher method. The results are shown in Table~\ref{voc_results}.
 CycleBNN secures a mean Average Precision (mAP) of 54.3, achieving the second-highest rank among the compared models while also realizing a significant operation reduction of 96.09\%. 
\begin{table}[!t]
	\setlength{\tabcolsep}{3mm}
	\renewcommand\arraystretch{0.9}
 \setlength{\tabcolsep}{2pt}
	\begin{center}
 \caption{The comparison of different methods on PASCAL VOC for object detection. W/A denotes the bit width of weights and activations. FLOPS denote the floating point precision operation in inference. MACS Pretraining- Training indicates The MACs used during pretraining on ImageNet an during the training on PASCAL VOC.  In this table, the best values in each column are highlighted in bold, and the second-best values are italicized.  In this table, the best values in each column are highlighted in bold, and the second-best values are italicized.}
	\label{voc_results}
\begin{tabular}{lccccc} 

\toprule
Methods & W/A & Params (M) & FLOPs (M) & mAP & \multicolumn{1}{r}{MACs Pretraining-Training} \\ 
\midrule
Full-precision & 32/32 & 100.28 & 31750 & 72.4 & 2.74E+16~\textbar{} 1.15E+9 \\
TWN & 2/32 & 24.54 & 8531 & 67.8 & - \\
DoReFa & 4/4 & 29.58 & 4661 & 69.2 & - \\ 
\midrule
BNN & 1/1 & 22.06 & 1275 & 42.0 & 2.74E+16~\textbar{} 1.15E+9 \\
XNOR-Net & 1/1 & 22.16 & 1279 & 50.2 & 2.74E+16~\textbar{} 1.15E+9 \\
BiDet & 1/1 & 22.06 & 1275 & 52.4 & 2.74E+16~\textbar{}  1.15E+9 \\
AutoBiDet & 1/1 & 22.06 & 1275 & 53.5 & 2.74E+16~\textbar{}  1.15E+9 \\
AdaBin & 1/1 & 22.47 & 1280 & \textbf{64.0} & 2.74E+16~\textbar{} 1.16E+9 \\
CycleBNN (Ours) & 1/1 & 22.06 & 1275 & \textit{54.3} & \textbf{1.05E+15~\textbar{} 4.42E+7} \\ 

\cmidrule(lr){2-6}

\multicolumn{1}{r}{Efficiency improv.} & \multicolumn{1}{l}{} & \multicolumn{1}{l}{} & \multicolumn{1}{l}{} &  & -96.09\% \\
\bottomrule
\end{tabular}
	\end{center}
	
\end{table}

\noindent\textbf{Considerations}
These achievements highlight CycleBNN's efficiency, with a remarkable reduction of 88.88\% MACs during training for the smaller dataset and 96.09\% on others, while still securing a position as the second or third best among the other methodologies. This showcases CycleBNN's competitive accuracy and training efficiency, making a compelling case for its adoption in resource-constrained environments without sacrificing accuracy.

\subsection{Ablation Study}
\begin{table} [!t]
\caption{The table below summarizes the results of the ablation study, showcasing the effects of varying normalization methods, activation functions, batch sizes, precision cycles, and scheduling on the CycleBNN model's performance on the CIFAR-10 classification benchmark. Each modification's impact is quantitatively assessed.}

\label{ablation_cycle}
\begin{center}
\begin{tabular}{l|l|llc|c} 
\toprule
Ablation                                                                       & Variant                                                                                                                                                                           & Params(M)              & Macs(G)                & Top-1(\%) &                            \\ 
\hline
Baseline                                                                       & \begin{tabular}[c]{@{}l@{}}Architecture: Resnet-18,\\Normalization: Z-Norm,\\Activation: Hardtanh,\\Precision cycles: 1,\\Batch size: 64,\\Precision scheduling: 2-8\end{tabular} & 11.69                  & 0.031                  & 92.93     &                            \\ 
\hline
Normalization                                                                  & Z-Norm $\rightarrow$ standardization                                                                                                                                                   & 11.69                  & 0.031                  & 93.18     & +0.25                      \\ 
\hline
\multirow{2}{*}{\begin{tabular}[c]{@{}l@{}}Activation\\function~\end{tabular}} & Hardtanh $\rightarrow$ Maxout                                                                                                                                                   & 11.98                  & 0.0314                 & 90.07     & -2.86                      \\
                                                                               & Hardtanh $\rightarrow$ ReLU                                                                                                                                                     & 11.69                  & 0.031                  & 88.36     & -4.57                      \\ 
\hline
\multirow{3}{*}{\begin{tabular}[c]{@{}l@{}}Batch\\size\end{tabular}}           & 64 $\rightarrow$ 8                                                                                                                                                              & \multirow{3}{*}{11.69} & \multirow{3}{*}{0.031} & 92.9      & -0.03                      \\
                                                                               & 64 $\rightarrow$ 16                                                                                                                                                             &                        &                        & 93.68     & +0.75                      \\
                                                                               & 64 $\rightarrow$ 32                                                                                                                                                             &                        &                        & 93.45     & +0.52                      \\ 
\hline
\multirow{4}{*}{\begin{tabular}[c]{@{}l@{}}Precision\\cycles\end{tabular}}     & 1 $\rightarrow$ 2                                                                                                                                                               & \multirow{4}{*}{11.69} & \multirow{4}{*}{0.031} & 92.89     & -0.04                      \\
                                                                               & 1 $\rightarrow$ 4                                                                                                                                                               &                        &                        & 92.68     & -0.25                      \\
                                                                               & 1 $\rightarrow$ 8                                                                                                                                                               &                        &                        & 92.98     & +0.05                      \\
                                                                               & 1~$\rightarrow$ 32                                                                                                                                                              &                        &                        & 92.74     & \multicolumn{1}{l}{-0.19}  \\ 
\hline
\multirow{4}{*}{\begin{tabular}[c]{@{}l@{}}Precision\\scheduling\end{tabular}} & 2-8 $\rightarrow$ 3-8                                                                                                                                                           & \multirow{4}{*}{11.69} & \multirow{4}{*}{0.031} & 93.15     & +0.27                      \\
                                                                               & 2-8 $\rightarrow$ 4-8                                                                                                                                                           &                        &                        & 93.04     & +0.11                      \\
                                                                               & 2-8 $\rightarrow$ 2-7                                                                                                                                                           &                        &                        & 92.95     & +0.02                      \\
                                                                               & 2-8 $\rightarrow$ 2-6                                                                                                                                                           &                        &                        & 93.16     & +0.28                      \\
\bottomrule
\end{tabular}

\end{center}

\end{table}
We conducted a comprehensive ablation study, focusing on the CycleBNN model with a Resnet-18 architecture trained on the CIFAR-10 dataset using the AdamW optimizer. The learning rate is initially set to 0.001 and follows a Cosine Annealing schedule, adhering to the methodology proposed in IR-Net \cite{qin2020forward}.  The baseline configuration employs a Resnet-18 architecture with Z-Norm normalization, Hardtanh activation, a batch size of 64,
a single precision cycle, and precision scheduling from 2 to 8. Each variant in the study alters only a single parameter from this base configuration.

We investigated the impact of activation functions in the CycleBNN model, motivated by their varied utilization in state-of-the-art methodologies \cite{tu2022adabin,xu2021recu,kim2021improving}. Our focus was on Hardtanh, MaxOut, and ReLU. According to the data presented in Table \ref{ablation_cycle}, the choice of activation function significantly influences the network's performance, with Hardtanh emerging as the preferred option for this method.

We evaluated various normalization techniques on CycleBNN, referencing methodologies employed in the current state of the art \cite{xu2021recu,tu2022adabin,qin2020forward}. Our experiments included Z-Normalization and a variant of Z-normalization that does not involve subtracting the mean. As the data in Table \ref{ablation_cycle} indicated, omitting the mean subtraction from the weights resulted in a performance enhancement of 0.25\%.

Given the training instability of BNNs cited in \cite{zhu2019binary}, we examine the effect of varying batch sizes on model accuracy. Batch sizes of 8, 16, 32, and 64 are tested. As detailed in the results on Table \ref{ablation_cycle}, the configuration employing a batch size of 16 achieved an accuracy improvement of 0.75\%.

The study is expanded to explore precision cycles, evaluating possible variations of (2, 4, 6, 8). According to the data in Table \ref{ablation_cycle}, these parameters do not significantly influence accuracy, with only the configuration utilizing 8 precision scheduling achieving a marginal accuracy increase of 0.05\%.

Different bits precision scheduling configurations (3-8, 4-8, 5-8, 2-7, 2-6) were evaluated for their impact on model performance. The results, thoroughly documented in Table \ref{ablation_cycle}, reveal that the 3-8 and 2-6 configurations exhibit comparable performance levels. However, using fewer bits, the 2-6 configuration allows greater savings in Training MACs.

 Collectively, these elements contribute to a notable improvement in model accuracy and efficiency. Specifically, underscores the importance of activation functions in maintaining non-linearity while ensuring stability during training. The refinement in normalization by omitting mean subtraction from Z-Normalization emerges as a pivotal factor in boosting model performance, highlighting the impact of weight normalization techniques on BNNs. Lastly, the exploration of precision scheduling demonstrates that strategic reduction in precision can yield computational savings without significantly compromising accuracy. These insights validate our design choices and underscore the importance of each element in achieving the optimal balance between accuracy, efficiency, and computational cost in the development of binary neural networks.

\section{Conclusions}
This study introduces CycleBNN, a novel BNN training method incorporating cyclic precision scheduling. We have established a mathematical foundation for this training methodology and evaluated its efficacy in image classification and object detection. Our comprehensive ablation study underscores our dedication to open science, with all codes and components available for replicating our findings and facilitating further advancements in efficient deep learning.
 CycleBNN realizes a significant decrease in MACs by 96.09 \% during ImageNet training sessions, 88.88\% reduction on CIFAR-10, and 96.09 \% on PASCAL-VOC. Additionally, CycleBNN can be optimized using XNOR and BitCount operations, like other BNNs, leading to a theoretical 58x speed-up and 32x memory reduction during inference \cite{rastegari2016xnor}. It notably enhances training efficiency, making it ideal for scenarios where energy, latency, or memory constraints are critical.
The scope of this research is limited to Binarized Convolutional Neural Networks, with its efficacy validated exclusively within this type of network. Application of this method to alternative architectures might necessitate additional investigation.
Future efforts can explore precision scheduling tailored to each network layer to enhance efficiency and accuracy. Expansion to architectures in \cite{guo2022join,le2023binaryvit} and application on specialized hardware \cite{huang2023precision,askarihemmat2023barvinn} supporting arbitrary bit precision.


\section*{Acknowledgements}
The research leading to these results has received funding from Project “Ecosistema dell’innovazione - Rome Technopole” financed by EU in NextGenerationEU plan through MUR Decree n. 1051 23.06.2022 - CUP B83C22002820006
%
%
\bibliographystyle{splncs04}
\bibliography{egbib}
\end{document}


\title{CycleBNN: Cyclic Precision Training in Binary Neural Networks} 

\titlerunning{CycleBNN: Cyclic Precision Training in Binary Neural Networks}

\author{Federico Fontana\inst{1}\orcidlink{0009-0007-0437-7832} \and
Romeo Lanzino\inst{1}\orcidlink{0000-0003-2939-3007} \and
Anxhelo Diko\inst{1}\orcidlink{0000-0002-3084-9377}
\and
Gian Luca Foresti\inst{2}\orcidlink{0000-0002-8425-6892}
\and
Luigi Cinque\inst{1}\orcidlink{0000-0001-9149-2175}
}

\authorrunning{F. Fontana et al.}

\institute{Sapienza University of Rome, Rome RM 00198, Italy \\ \email{ \{fontana.f, lanzino, diko, cinque\}@di.uniroma1.it} \and
UNIUD University of Udine,  Udine UD 33100, Italy \\
\email{ gianluca.foresti@uniud.it}}


\maketitle

\section*{Supplementary Material}

CycleBNN introduces a novel approach that combines binary neural networks (BNNs) with cyclic precision training to improve the efficiency of BNN training while maintaining competitive performance levels. This method dynamically adjusts the precision during training phases, balancing training efficiency and model performance.
\section{Elaboration on Training, Testing, and Code Availability Procedures}

As explicitly stated within the primary manuscript, we will commit to releasing both the source code and the trained model weights to the public domain after formally accepting our paper. Currently, the source code is provisionally available to reviewers to facilitate the review process. Access can be obtained through the URL: \url{https://github.com/fedeloper/CycleBNN/}.

The provided README file contains detailed instructions to ensure our research findings' reproducibility. It includes guidelines for configuring a computational environment using Docker and establishing a consistent standard for reproducing our results. Furthermore, the README elaborates on the specific procedures for training our models on ImageNet and CIFAR-10. It also details the methodology for fine-tuning these pre-trained models and testing them on the Pascal VOC dataset. 

\section{Math Considerations on Low Precision and Quantization Error}
This study investigates the effect of precision level $p$ on the Quantization Error (\textit{QE}) in BNNs. Our analysis begins by defining the \textit{QE} and static quantization formula. We then demonstrate that static quantization and \textit{Sign} formula are equivalent for a range of [-1,1] and a precision of 1 bit. Subsequently, we numerically approximate the expected \textit{QE} for various bit precisions and the actual weight distributions.

\subsection{Define Domain and Function}

This is the static quantization formula:

\begin{equation}
\textit{Q}(x,p,m,M) = m + \frac{\left\lfloor \frac{(x-m)(2^p-1)}{M-m}+\frac{1}{2} \right\rfloor (M-m)}{2^p -1}
\end{equation}
 where $x$ is the value, $p$ is the number of bits used, $m$ and $M$ are the minimum value of the quantization and the maximum value.

\begin{equation}
\operatorname{\textit{QE}(w^r)} = \int_{-\infty}^{+\infty}f(w^r)\big(w^r - \alpha \operatorname{\textit{Sign}}(w^r)\big)^2\mathrm{d}w^r
\end{equation}
\label{QE}
This is the formalization of the \textit{QE}, where $f$ is the Probability Density Function, $w^r$ are the full precision weights.
Applying the static quantization formula to \textit{QE}, we obtain the following:
\begin{equation}
{\operatorname{\textit{QE}(w^r,p,m,M)} = \int_{-\infty}^{+\infty} \textit{f}(\textit{Q}(w^r, p, m, M))\big(\textit{Q}(w^r, p, m, M) - \alpha \operatorname{\textit{Sign}}(Q(w^r, p, m, M))\big)^2\mathrm{d}w^r}
\end{equation}

\subsection{Demonstrate \( \mathit{Sign(x)}  = Q(x,1,-1,1) \)}

Intuitively, we can express the \textit{Sign} function using the precision simulation formula as follows, in the range [-1, 1] and with $x \neq 0$:

\begin{align}
    \nonumber \textit{Q}(x,p,m,M) = m + \frac{\left\lfloor \frac{(x-m)(2^p-1)}{M-m}+\frac{1}{2} \right\rfloor *(M-m)}{2^p -1} \Rightarrow \\
    \nonumber \Rightarrow \left\{  p= 1, m = -1, M = 1, -1 => x >= 1, x \neq  0\right\} \Rightarrow \\
    \nonumber  \Rightarrow \textit{Q}(x,1,1,-1) = 2\left\lfloor \frac{x}{2}  +1 \right\rfloor -1\Rightarrow 
     \\  \Rightarrow \textit{Q}(x,1,1,-1) = \begin{Bmatrix}
                                    x < 0 & -1 \\
                                     x > 0 & 1 \\
                                     x = 0  & undefined
                                    \end{Bmatrix} = Sign(x)
                                    \label{eq_sign}
  \end{align}

\subsection{Demonstrating that if $p=1$ then \textit{QE} is 0}

Starting from:
\begin{equation}
{\operatorname{\textit{QE}(w^r,p,m,M)} = \int_{-\infty}^{+\infty} f(\textit{Q}(w^r, p, m, M))\big(Q(w^r, p, m, M) - \alpha \operatorname{\textit{Sign}}(\textit{Q}(w^r, p, m, M))\big)^2\mathrm{d}w^r}
\end{equation}

Using eq. \ref{eq_sign} and with $p = 1$, $m = -1$ and $M = 1$ we get:

\begin{equation}
\operatorname{\textit{QE}(w^r,1,-1,1)} = \int_{-\infty}^{+\infty}f( \operatorname{\textit{Sign}}(w^r))\big( \operatorname{\textit{Sign}}(w^r) - \alpha \operatorname{\textit{Sign}}(w^r)\big)^2\mathrm{d}w^r
\end{equation}
with $\alpha = 1$ then $\textit{QE}(w^r,1,-1,1) = 0$, in the extreme case when the training was done with 1-bit precision, the \textit{QE} will be 0.
\subsection{Numerical approximation of the Integrals }
\begin{figure}[h]
    \centering
    \caption{In this figure are shown the weight distributions for each binarized convolutional layer. Figures of binarized convolutions are proceeded in alphabetical sequence, with each figure featuring a red line representing the Gaussian fit of the distribution, accompanied by descriptions outlining the fit's characteristics.}
    \label{gauss}
    \begin{subfigure}{.24\textwidth}
        \centering
        \includegraphics[width=.9\linewidth]{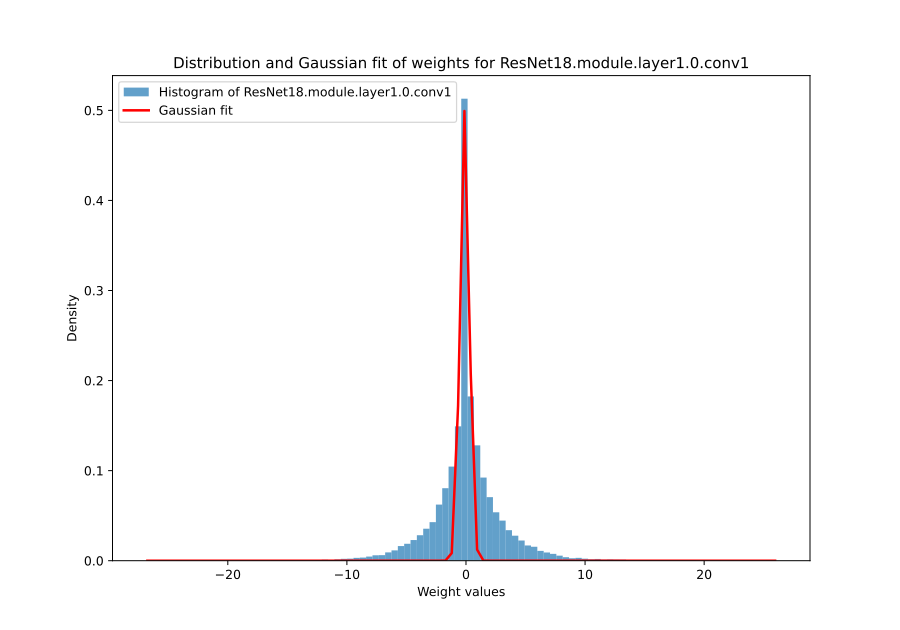}
        \caption{Amplitude: 0.50,\\ $\mu$: -0.11, $\sigma$: 0.38}
        
    \end{subfigure}%
    \begin{subfigure}{.24\textwidth}
        \centering
        \includegraphics[width=.9\linewidth]{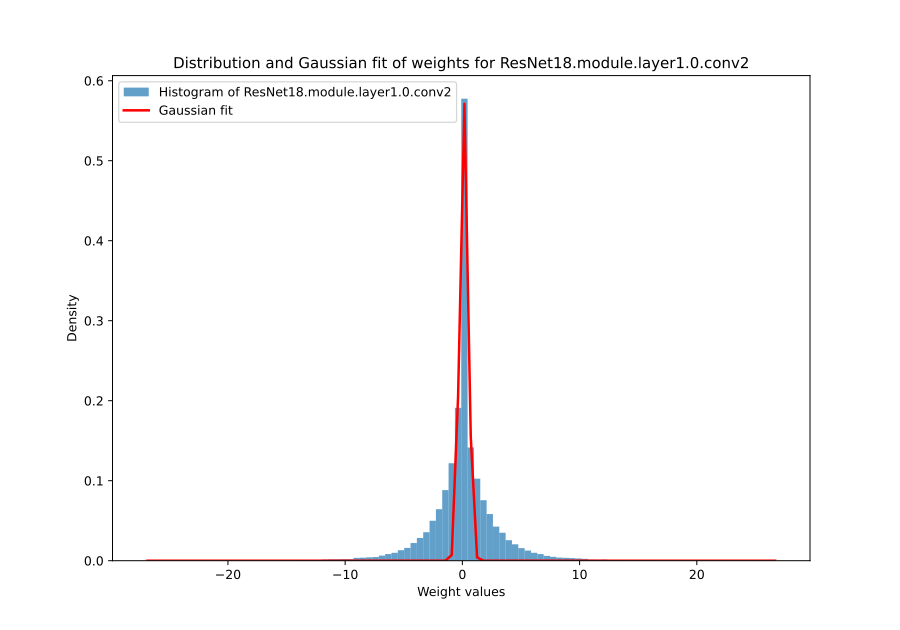}
        \caption{Amplitude: 0.57, \\$\mu$: 0.14, $\sigma$: 0.35}
       
    \end{subfigure}
    \begin{subfigure}{.24\textwidth}
        \centering
         \includegraphics[width=.9\linewidth]{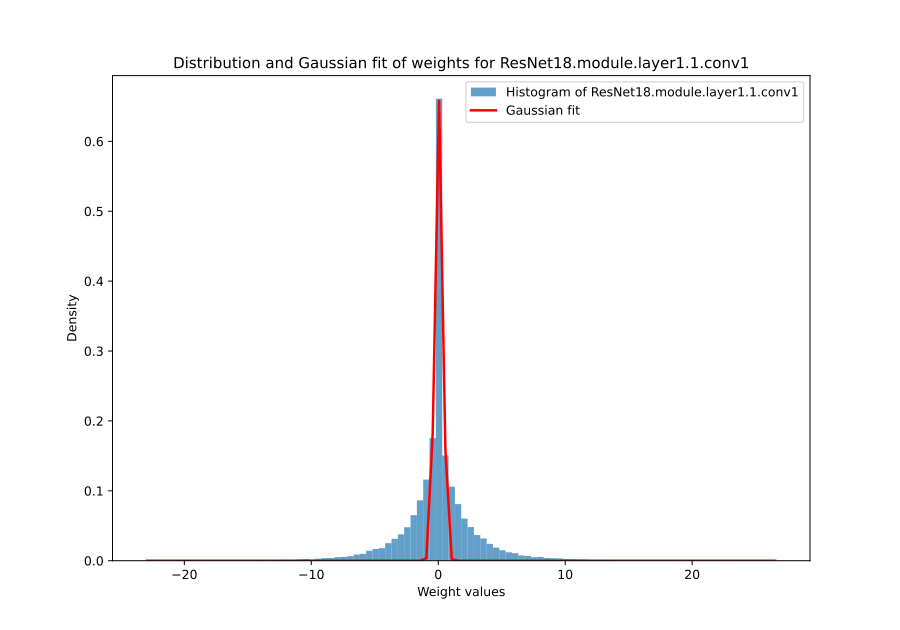}
        \caption{Amplitude: 0.66, \\$\mu$: 0.05, $\sigma$: 0.30}
        \label{fig:sub1-3}
    \end{subfigure}%
    \begin{subfigure}{.24\textwidth}
        \centering
         \includegraphics[width=.9\linewidth]{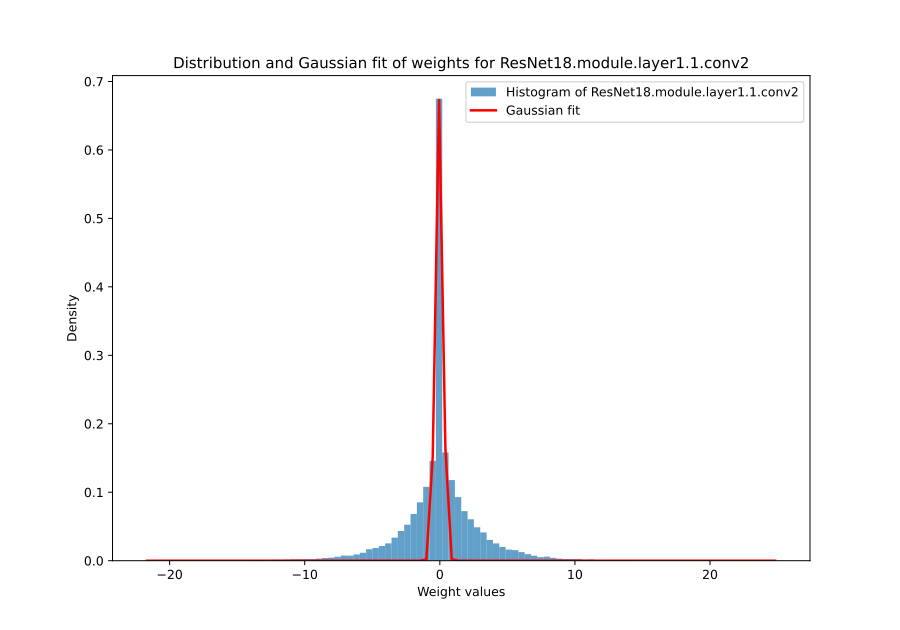}
        \caption{Amplitude: 0.67, \\$\mu$: -0.05, $\sigma$: 0.28}
      
    \end{subfigure}


    \begin{subfigure}{.24\textwidth}
        \centering
       \includegraphics[width=.9\linewidth]{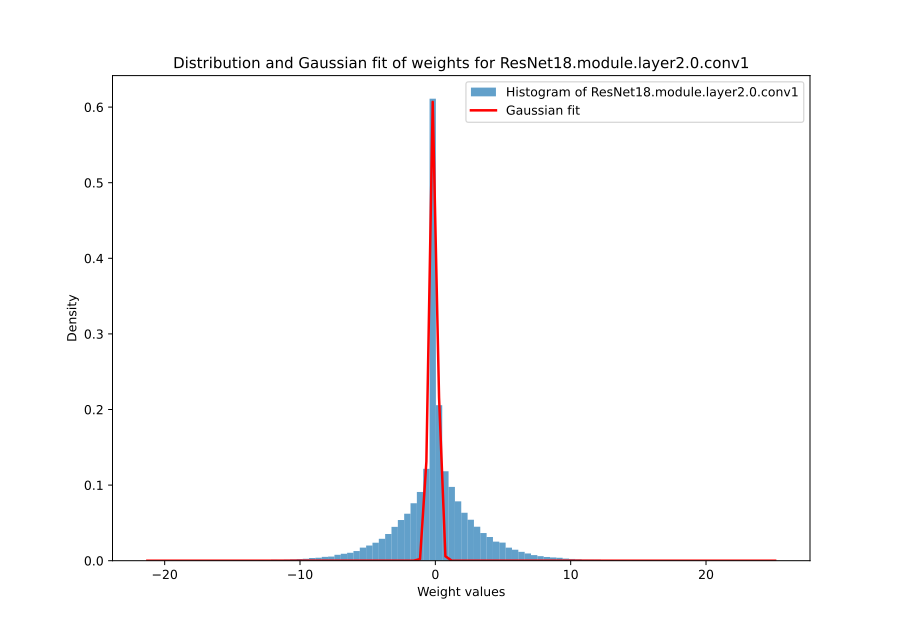}
        \caption{Amplitude: 0.61, \\$\mu$: -0.15, $\sigma$: 0.29}
        
    \end{subfigure}%
    \begin{subfigure}{.24\textwidth}
        \centering
        \includegraphics[width=.9\linewidth]{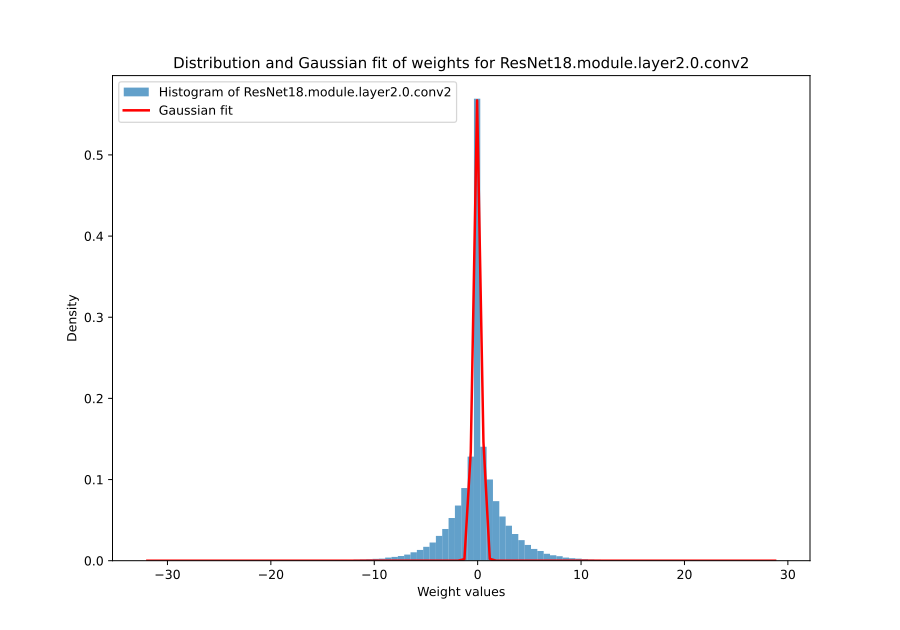}
        \caption{Amplitude: 0.57, \\$\mu$: -0.04, $\sigma$: 0.37}
        
    \end{subfigure}
    \begin{subfigure}{.24\textwidth}
        \centering
        \includegraphics[width=.9\linewidth]{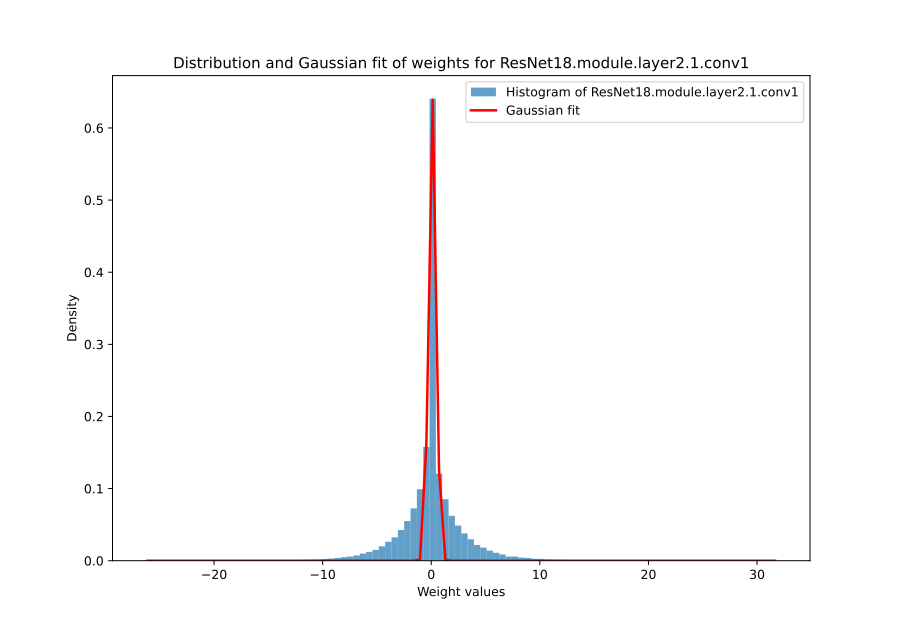}
        \caption{Amplitude: 0.64, \\$\mu$: 0.11, $\sigma$: 0.34}
       
    \end{subfigure}%
    \begin{subfigure}{.24\textwidth}
        \centering
         \includegraphics[width=.9\linewidth]{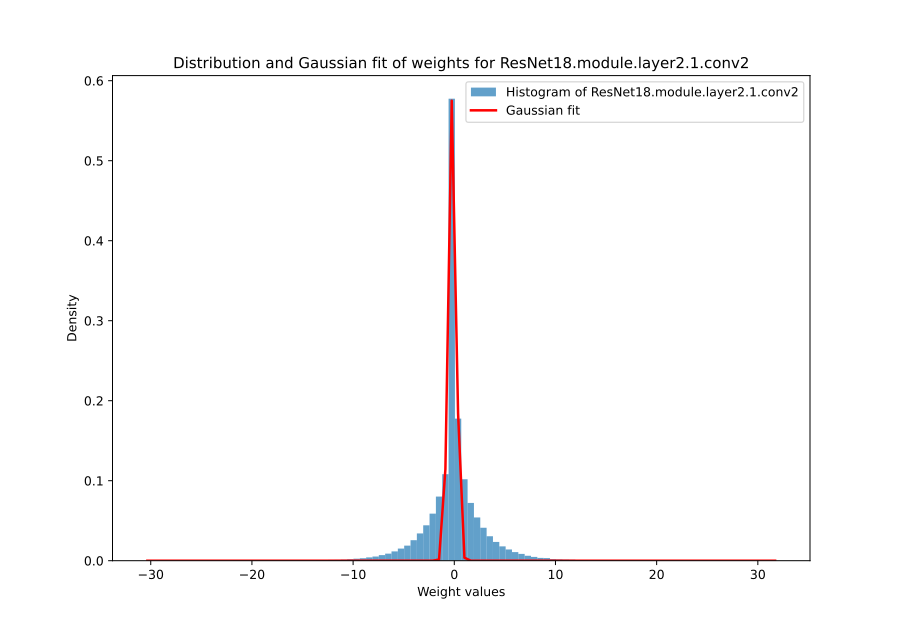}
        \caption{Amplitude: 0.58, \\$\mu$: -0.20, $\sigma$: 0.38}
      
    \end{subfigure}

 \begin{subfigure}{.24\textwidth}
        \centering
          \includegraphics[width=.9\linewidth]{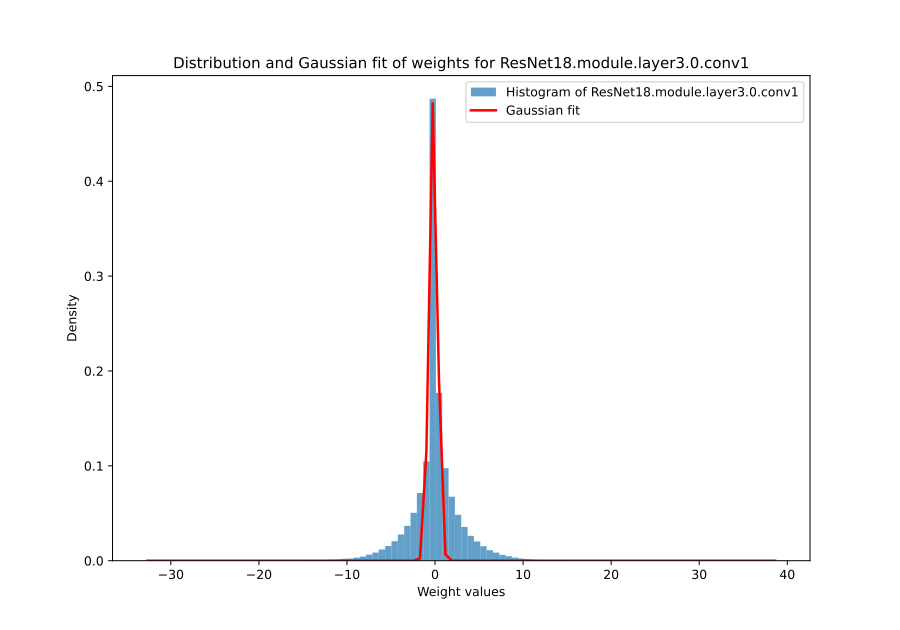}
        \caption{Amplitude: 0.49, \\$\mu$: -0.19, $\sigma$: 0.47}
        
    \end{subfigure}%
    \begin{subfigure}{.24\textwidth}
        \centering
        \includegraphics[width=.9\linewidth]{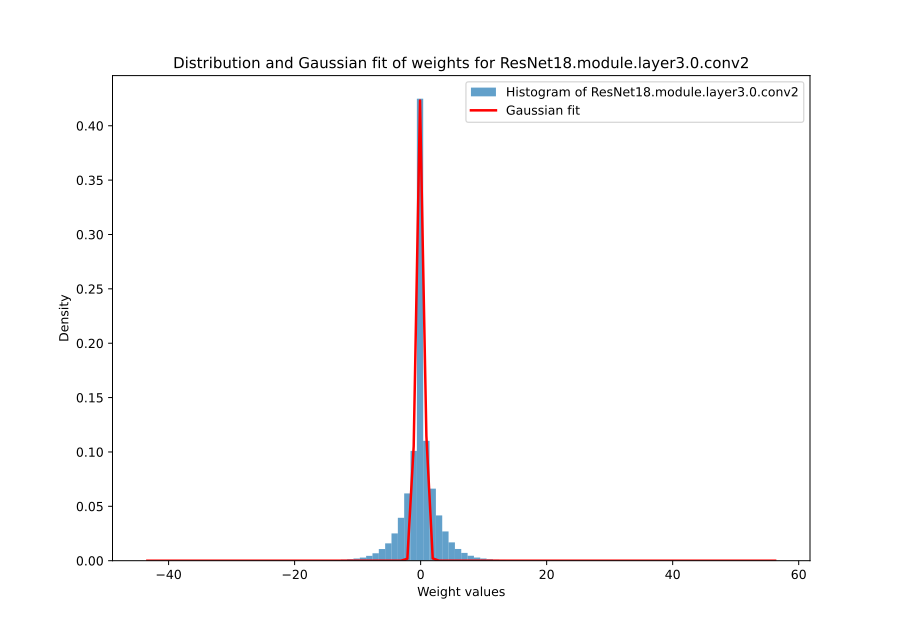}
        \caption{Amplitude: 0.42, \\$\mu$: -0.08, $\sigma$: 0.61}
        
    \end{subfigure}
    \begin{subfigure}{.24\textwidth}
        \centering
        \includegraphics[width=.9\linewidth]{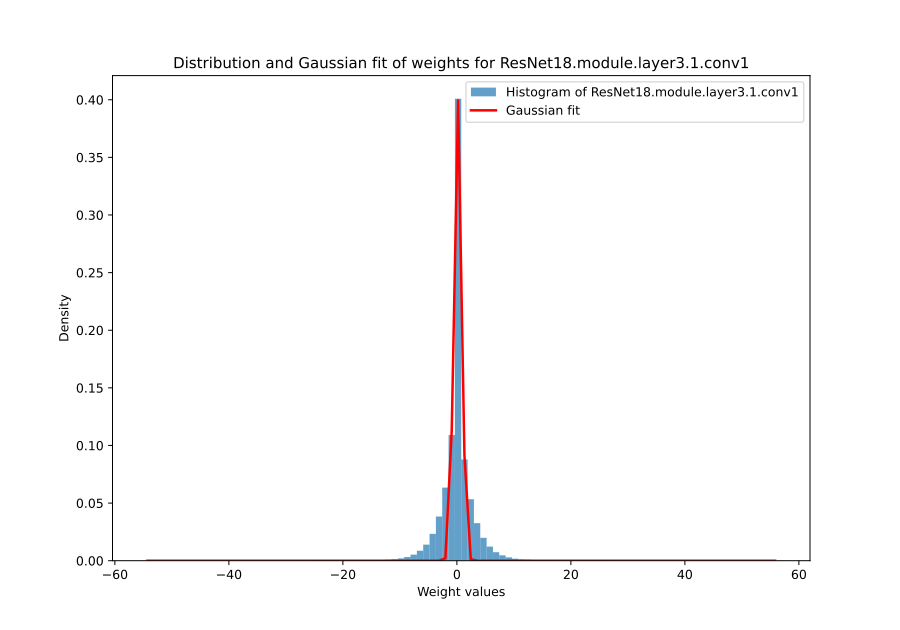}
        \caption{Amplitude: 0.40, \\$\mu$: 0.17, $\sigma$: 0.67}
       
    \end{subfigure}%
    \begin{subfigure}{.24\textwidth}
        \centering
         \includegraphics[width=.9\linewidth]{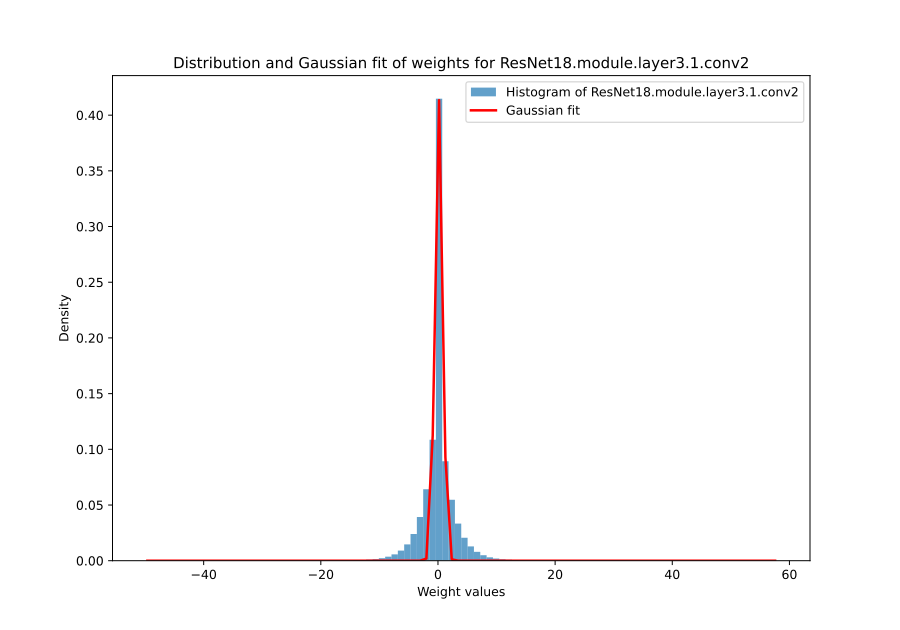}
        \caption{Amplitude: 0.41, \\$\mu$: 0.15, $\sigma$: 0.65}
      
    \end{subfigure}

 \begin{subfigure}{.24\textwidth}
        \centering
          \includegraphics[width=.9\linewidth]{distributions/ResNet18.module.layer3.0.conv1.png}
        \caption{Amplitude: 0.42, \\$\mu$: 0.13, $\sigma$: 0.60}

    \end{subfigure}%
    \begin{subfigure}{.24\textwidth}
        \centering
        \includegraphics[width=.9\linewidth]{distributions/ResNet18.module.layer3.0.conv2.png}
        \caption{Amplitude: 0.35, \\$\mu$: 0.09, $\sigma$: 0.70}
        
    \end{subfigure}
    \begin{subfigure}{.24\textwidth}
        \centering
       \includegraphics[width=.9\linewidth]{distributions/ResNet18.module.layer3.1.conv1.png}
        \caption{Amplitude: 0.40, \\$\mu$: 0.14, $\sigma$: 0.57}
       
    \end{subfigure}%
    \begin{subfigure}{.24\textwidth}
        \centering
        \includegraphics[width=.9\linewidth]{distributions/ResNet18.module.layer3.1.conv2.png}
        \caption{Amplitude: 0.39, \\$\mu$: -0.02, $\sigma$: 0.59}
      
    \end{subfigure}

\end{figure}

Numerical evidence is presented to substantiate the hypothesis that training with reduced precision reduces quantization error. For each binarized convolution in the trained ResNet-18, \textit{QE} integrals were calculated at various bit precisions, particularly within the \([-15, 15]\) interval. This interval has been empirically identified as the maximal weight range in the trained network. A graphical representation of the distribution of each binarized convolutional layer is shown in Fig. \ref{gauss}, then the numerical approximations of \textit{QE}s are shown in Table~\ref{approx}.  
\begin{table}[]
        \caption{This table shows the numerical approximation of the integral in Eq. \ref{eq_sign} for varying bit precision and each of the binarized convolutional layers in ResNet-18, and probability density functions. The letters on columns correspond to the letters on Fig. \ref{gauss} }
\resizebox{\textwidth}{!}{%
\begin{tabular}{ccccccccccccccccc} 
\toprule
Bits & a & b & c & d & e & f & g & h & i & j & k & l & m & n & o & p\\  
\cmidrule{1-17}
2 & 0.192 & 0.205 & 0.214 & 0.205 & 0.189 & 0.218 & 0.227 & 0.216 & 0.210 & 0.216 & 0.194 & 0.197 & 0.210 & 0.216 & 0.194 & 0.197 \\
3 & 0.240 & 0.259 & 0.287 & 0.280 & 0.247 & 0.276 & 0.291 & 0.264 & 0.249 & 0.248 & 0.224 & 0.228 & 0.249 & 0.248 & 0.224 & 0.228 \\
4 & 0.248 & 0.267 & 0.298 & 0.292 & 0.256 & 0.285 & 0.301 & 0.272 & 0.256 & 0.253 & 0.230 & 0.233 & 0.256 & 0.253 & 0.229 & 0.233 \\
5 & 0.249 & 0.269 & 0.301 & 0.294 & 0.258 & 0.287 & 0.303 & 0.273 & 0.257 & 0.254 & 0.231 & 0.234 & 0.257 & 0.254 & 0.231 & 0.234 \\
6 & 0.250 & 0.270 & 0.301 & 0.295 & 0.258 & 0.288 & 0.304 & 0.274 & 0.258 & 0.255 & 0.231 & 0.234 & 0.258 & 0.255 & 0.231 & 0.234 \\
7 & 0.250 & 0.270 & 0.301 & 0.295 & 0.259 & 0.288 & 0.304 & 0.274 & 0.258 & 0.255 & 0.231 & 0.234 & 0.258 & 0.255 & 0.231 & 0.234 \\
8 & 0.250 & 0.270 & 0.301 & 0.295 & 0.259 & 0.288 & 0.304 & 0.274 & 0.258 & 0.255 & 0.231 & 0.234 & 0.258 & 0.255 & 0.231 & 0.234 \\
9 & 0.250 & 0.270 & 0.301 & 0.295 & 0.259 & 0.288 & 0.304 & 0.274 & 0.258 & 0.255 & 0.231 & 0.234 & 0.258 & 0.255 & 0.231 & 0.234 \\
10 & 0.250 & 0.270 & 0.301 & 0.295 & 0.259 & 0.288 & 0.304 & 0.274 & 0.258 & 0.255 & 0.231 & 0.234 & 0.258 & 0.255 & 0.231 & 0.235 \\
11 & 0.250 & 0.270 & 0.301 & 0.295 & 0.259 & 0.288 & 0.304 & 0.274 & 0.258 & 0.255 & 0.231 & 0.234 & 0.258 & 0.255 & 0.231 & 0.234 \\
12 & 0.250 & 0.270 & 0.301 & 0.295 & 0.259 & 0.288 & 0.304 & 0.274 & 0.258 & 0.255 & 0.231 & 0.234 & 0.258 & 0.255 & 0.231 & 0.234 \\
\bottomrule
\label{approx}
\end{tabular}}

\end{table}
\subsection{Considerations on Quantization Error}
As shown in Table~\ref{approx}, the numerical evaluation of integrals corroborates our hypothesis that the binarization of quantized networks results in reduced quantization error (Eq.~\ref{QE}). This is particularly evident when considering precision bits ranging from 2 to 5.